\documentclass[preprint,12pt]{elsarticle}
\usepackage{tikz}
\usetikzlibrary{shapes.geometric,positioning}
\usepackage{bm}
\usepackage{amsmath}
\usepackage{hyperref}

\hypersetup{
  colorlinks   = true, 
  urlcolor     = blue, 
  linkcolor    = blue, 
  citecolor   = red 
}

\usepackage{amssymb}

\usepackage{lineno}

\journal{arXiv}

\begin{document}

\begin{frontmatter}



\title{Machine Learning of Linear Differential Equations using Gaussian Processes}


\author{Maziar Raissi$^{1}$ and George Em. Karniadakis$^{1}$}

\address{$^{1}$Division of Applied Mathematics, Brown University,\\ 182 George Street, Providence, RI 02912}

\begin{abstract}
This work leverages recent advances in probabilistic machine learning to discover conservation laws expressed by parametric linear equations. Such equations involve, but are not limited to, ordinary and partial differential, integro-differential, and fractional order operators. Here, Gaussian process priors are modified according to the particular form of such operators and are employed to infer parameters of the linear equations from scarce and possibly noisy observations. Such observations may come from experiments or ``black-box" computer simulations.
\end{abstract}


\end{frontmatter}


\section{Introduction} A grand challenge with great opportunities facing researchers is to develop a coherent framework that enables scientists to blend conservation laws expressed by differential equations with the vast data sets available in many fields of engineering, science and technology. In particular, this article investigates conservation laws of the form

\begin{center}
\begin{tikzpicture}[>=latex, baseline=(current  bounding  box.center)]
  \node [draw,scale=1.25,fill,fill opacity=0.1,text opacity=1] (L) {$\mathcal{L}_x^\phi:$ $\phi {\ =\ } ?$};
  \node [left=of L] (input) {$u(x)$};
  \node [right=of L] (output) {$f(x)$,};

  \draw [->, line width=0.5mm] (input) -- (L);
  \draw [->, line width=0.5mm] (L) -- (output);
\end{tikzpicture}
\end{center}

\noindent which model the relationship between two black-box functions $u(x)$ and $f(x)$. Here, 
\begin{equation}\label{eq:key}
f(x) = \mathcal{L}_x^\phi u(x)
\end{equation}
and $\mathcal{L}_x^\phi$ is a parametric linear operator identified by its parameters $\phi$. Given noisy observations $\{\bm{X}_u,\bm{y}_u\}$, $\{\bm{X}_f,\bm{y}_f\}$ of $u(x)$, $f(x)$, respectively, the aim is to learn the parameters $\phi$ and hence the governing equation which best describes the data. Such problems are ubiquitous in science, and in mathematical literature are often referred to as ``inverse problems'' (see \cite{kaipio2006statistical, stuart2010inverse}). To provide a unified framework for resolving such problems, this work employs and modifies Gaussian processes (GPs) (see \cite{williams2006gaussian, murphy2012machine}), which is a non-parametric Bayesian machine learning technique. Quoting Diaconis \cite{diaconis1988bayesian}, ``once we allow that we don't know $f$ (and $u$), but do know some things, it becomes natural to take a Bayesian approach". The Bayesian procedure adopted here, namely Gaussian processes,  provides a flexible prior distribution over functions, enjoys analytical tractability, and has a fully probabilistic work-flow that returns robust posterior variance estimates which quantify uncertainty in a natural way. Moreover, Gaussian processes are among a class of methods known as kernel machines (see \cite{vapnik2013nature, scholkopf2002learning, tipping2001sparse}) and have analogies with regularization approaches (see \cite{tikhonov1963solution, Tikhonov/Arsenin/77, poggio1990networks}). However, they are distinguished by their probabilistic viewpoint and their powerful traning procedure. Along exactly the same lines, the methodology outlined in this work is related to and yet fundamentally differentiated from the so-called ``meshless" methods (see \cite{franke1998solving, fasshauer2013kernel, owhadi2015bayesian, cockayne2016probabilistic}) for differential equations. Furthermore, differential equations are the cornerstone of a diverse range of applied sciences and engineering fields. However, use within statistics and machine learning, and combination with probabilistic models is less explored. Perhaps the most significant related work in this direction is latent force models \cite{alvarez2013linear, alvarez2009latent}. Such models generalize latent variable models \cite{lawrence2005probabilistic,lawrence2004gaussian, titsias2010bayesian} using differential equations. In sharp contrast to latent force models, this work bypasses the need for solving differential equations either analytically or numerically by placing the Gaussian process prior on $u(x)$ rather than on $f(x)$. Additionally, equation \ref{eq:key} can be further motivated by considering the familiar cases where 
\[
\mathcal{L}_x^\phi u(x) = (\kappa * u)(x) = \int \kappa(x-x';\phi)u(x')dx',
\]
for some kernel $\kappa$. This particular instance corresponds to the well-known convolved Gaussian processes \cite{higdon2002space,boyle2004dependent} which are suitable for multi-output purposes \cite{alvarez2009sparse}. Moreover, yet another special and interesting case arises by assuming $\kappa(x-x';\phi) = \phi \delta(x-x')$, with $\delta$ being the Kronecker delta, which yields
\[
\mathcal{L}_x^\phi u(x) = \phi u(x).
\]
This results in the so-called recursive co-kriging \cite{kennedy2000predicting,le2014recursive} model $f(x) = \phi u(x) + v(x)$. Here, $v(x)$ is a latent function and is included to capture unexplained factors. Recursive co-kriging models can be employed to create a platform for blending multiple information sources of variable fidelity, e.g., experimental data, high-fidelity numerical simulations, expert opinion, etc. The main assumption in multi-fidelity settings is that the data on the high-fidelity function $f(x)$ are scarce since they are generated by an accurate but costly process, while the data on the low fidelity function $u(x)$ are less accurate, cheap to generate, and hence abundant.

\section{Methodology} The proposed data-driven algorithm for learning general parametric linear equations of the form \ref{eq:key} employs Gaussian process priors that are tailored to the corresponding differential operators.

\subsection{Prior} Specifically, the algorithm starts by making the assumption that $u(x)$ is Gaussian process \cite{williams2006gaussian} with mean $0$ and covariance function $k_{uu}(x,x';\theta)$, i.e.,
\[
u(x) \sim \mathcal{GP}(0, k_{uu}(x,x';\theta)),
\]
where $\theta$ denotes the hyper-parameters of the kernel $k_{uu}$. The key observation to make is that any linear transformation of a Gaussian process such as differentiation and integration is still a Gaussian process. Consequently,
\[
\mathcal{L}_x^\phi u(x) = f(x) \sim \mathcal{GP}(0, k_{ff}(x,x';\theta,\phi)),
\]
with the following fundamental relationship between the kernels $k_{uu}$ and $k_{ff}$,
\begin{equation}\label{eq:kernelk}
k_{ff}(x,x';\theta,\phi) = \mathcal{L}_x^\phi \mathcal{L}^\phi_{x'} k_{uu}(x,x';\theta).
\end{equation}
Moreover, the covariance between $u(x)$ and $f(x')$, and similarly the one between $f(x)$ and $u(x')$, are given by $k_{uf}(x,x';\theta,\phi) = \mathcal{L}^\phi_{x'} k_{uu}(x,x';\theta)$, and $k_{fu}(x,x';\theta,\phi) = \mathcal{L}^\phi_{x} k_{uu}(x,x';\theta)$, respectively. The main contribution of this work can be best recognized by noticing how the parameters $\phi$ of the operator $\mathcal{L}_x^\phi$ are turned into hyper-paramters of the kernels $k_{ff}, k_{uf},$ and $k_{fu}$.

\subsubsection*{Kernels \cite{williams2006gaussian}} Without loss of generality, all Gaussian process priors used in this work are assumed to have a squared exponential covariance function, i.e.,
\[
k_{uu}(x,x';\theta) = \sigma_{u}^2 \exp\left(-\frac12\sum_{d=1}^D w_{d}(x_d - x'_d)^2\right),
\]
where $\sigma_{u}^2$ is a variance parameter, $x$ is a $D$-dimensional vector that includes spatial or temporal coordinates, and $\theta = \left(\sigma_{u}^{2},(w_{d})_{d=1}^D\right)$. Moreover, anisotropy across input dimensions is handled by Automatic Relevance Determination (ARD) weights 
$w_{d}$. From a theoretical point of view, each kernel gives rise to a Reproducing Kernel Hilbert Space \cite{aronszajn1950theory, saitoh1988theory, berlinet2011reproducing} that defines a class of functions that can be represented by this kernel. In particular, the squared exponential covariance function chosen above, implies smooth approximations. More complex function classes can be accommodated by appropriately choosing kernels.

\subsection{Training}
The hyper-parameters $\theta$ and more importantly the parameters $\phi$ of the linear operator $\mathcal{L}_x^\phi$ can be trained by employing a Quasi-Newton optimizer L-BFGS \cite{liu1989limited} to minimize the negative log marginal likelihood \cite{williams2006gaussian}
\begin{equation}
\mathcal{NLML}(\phi,\theta,\sigma_{n_u}^2, \sigma_{n_f}^2) = -\log p(\bm{y}| \phi, \theta, \sigma_{n_u}^2, \sigma_{n_f}^2),
\end{equation}
where $\bm{y} = \left[\begin{array}{c}
\bm{y}_u \\ 
\bm{y}_f
\end{array} \right]$, $p(\bm{y} | \phi, \theta,\sigma_{n_u}^2,\sigma_{n_f}^2) = \mathcal{N}\left(\bm{0}, \bm{K}\right)$, and $\bm{K}$ is given by
\begin{equation}
\bm{K} = \left[ \begin{array}{cc}
k_{uu}(\bm{X}_u,\bm{X}_u;\theta) + \sigma_{n_u}^2 \bm{I}_{n_u} & k_{uf}(\bm{X}_u,\bm{X}_f;\theta,\phi)\\
k_{fu}(\bm{X}_f,\bm{X}_u;\theta,\phi) & k_{ff}(\bm{X}_f,\bm{X}_f;\theta,\phi) + \sigma_{n_f}^2 \bm{I}_{n_f}
\end{array}  \right].
\end{equation}
Here, $\sigma_{n_u}^2$ and $\sigma_{n_f}^2$ are included to capture noise in the data. The implicit underlying assumption is that $\bm{y}_u = u(\bm{X}_u) + \bm{\epsilon}_u$, $\bm{y}_f = f(\bm{X}_f) + \bm{\epsilon}_f$ with $\bm{\epsilon}_u \sim \mathcal{N}(\bm{0},\sigma_{n_u}^2 \bm{I}_{n_u})$, $\bm{\epsilon}_f \sim \mathcal{N}(\bm{0},\sigma_{n_f}^2 \bm{I}_{n_f})$. Moreover, $\bm{\epsilon}_u$ and $\bm{\epsilon}_f$ are assumed to be independent. It is worth mentioning that the marginal likelihood does not simply favor the models that fit the training data best. In fact, it induces an automatic trade-off between data-fit and model complexity. This effect is called Occam's razor \cite{rasmussen2001occam} after William of Occam 1285--1349 who encouraged simplicity in explanations by the principle: ``plurality should not be assumed without necessity". The flexible training procedure outlined above distinguishes Gaussian processes and consequently this work from other kernel-based methods. The most computationally intensive part of training is associated with inverting dense covariance matrices $\bm{K}$. This scales cubically with the number of training data in $\bm{y}$. Although this scaling is a well-known limitation of Gaussian process regression, it must be emphasize that it has been effectively addressed by the recent works of \cite{snelson2005sparse, hensman2013gaussian}.

\subsection{Prediction} Having trained the model, one can predict the values $u(x)$ and $f(x)$ at a new test point $x$ by writing the posterior distributions
\begin{eqnarray}
p(u(x) | \bm{y}) &=& \mathcal{N}\left(\overline{u}(x), s_u^2(x)\right),\\
p(f(x) | \bm{y}) &=& \mathcal{N}\left(\overline{f}(x), s_f^2(x)\right),
\end{eqnarray}
with
\begin{eqnarray}
\overline{u}(x) = \bm{q}_u^T \bm{K}^{-1} \bm{y},\ s^2_u(x) = k_{uu}(x,x) - \bm{q}_u^T \bm{K}^{-1} \bm{q}_u,\ \bm{q}_u^T = \left[k_{uu}(x,\bm{X}_u) \ k_{uf}(x,\bm{X}_f)\right],\nonumber\\
\overline{f}(x) = \bm{q}_f^T \bm{K}^{-1} \bm{y},\ s^2_f(x) = k_{ff}(x,x) - \bm{q}_f^T \bm{K}^{-1} \bm{q}_f,\ \bm{q}_f^T = \left[k_{fu}(x,\bm{X}_u)\ k_{ff}(x,\bm{X}_f)\right],\nonumber
\end{eqnarray}
where for notational convenience the dependence of kernels on hyper-parameters and parameters is dropped. The posterior variances $s_u^2(x)$ and $s_f^2(x)$ can be used as good indicators of how confident one could be about the estimated parameters $\phi$ of the linear operator $\mathcal{L}_x^\phi$ and predictions made based on these parameters. Furthermore, such built-in quantification of uncertainty encoded in the posterior variances is a direct consequence of the Bayesian approach adopted in this work. Although not pursued here, this information is very useful in designing a data acquisition plan, often referred to as {\em active learning} \cite{cohn1996active, krause2007nonmyopic, mackay1992information}, that can be used to optimally enhance our knowledge about the parametric linear equation under consideration.

\section{Results} The proposed algorithm provides an entirely agnostic treatment of linear operators, which can be of fundamentally different nature. For example, one can seamlessly learn parametric integro-differential, time-dependent, or even fractional equations. This generality will be demonstrated using three benchmark problems with simulated data. Moreover, the methodology will be applied to a fundamental problem in functional genomics, namely determining the structure and dynamics of genetic networks based on real expression data \cite{perkins2006reverse} on early \emph{Drosophila melanogaster} development.

\begin{figure}
\centering
\includegraphics[width=0.9\textwidth]{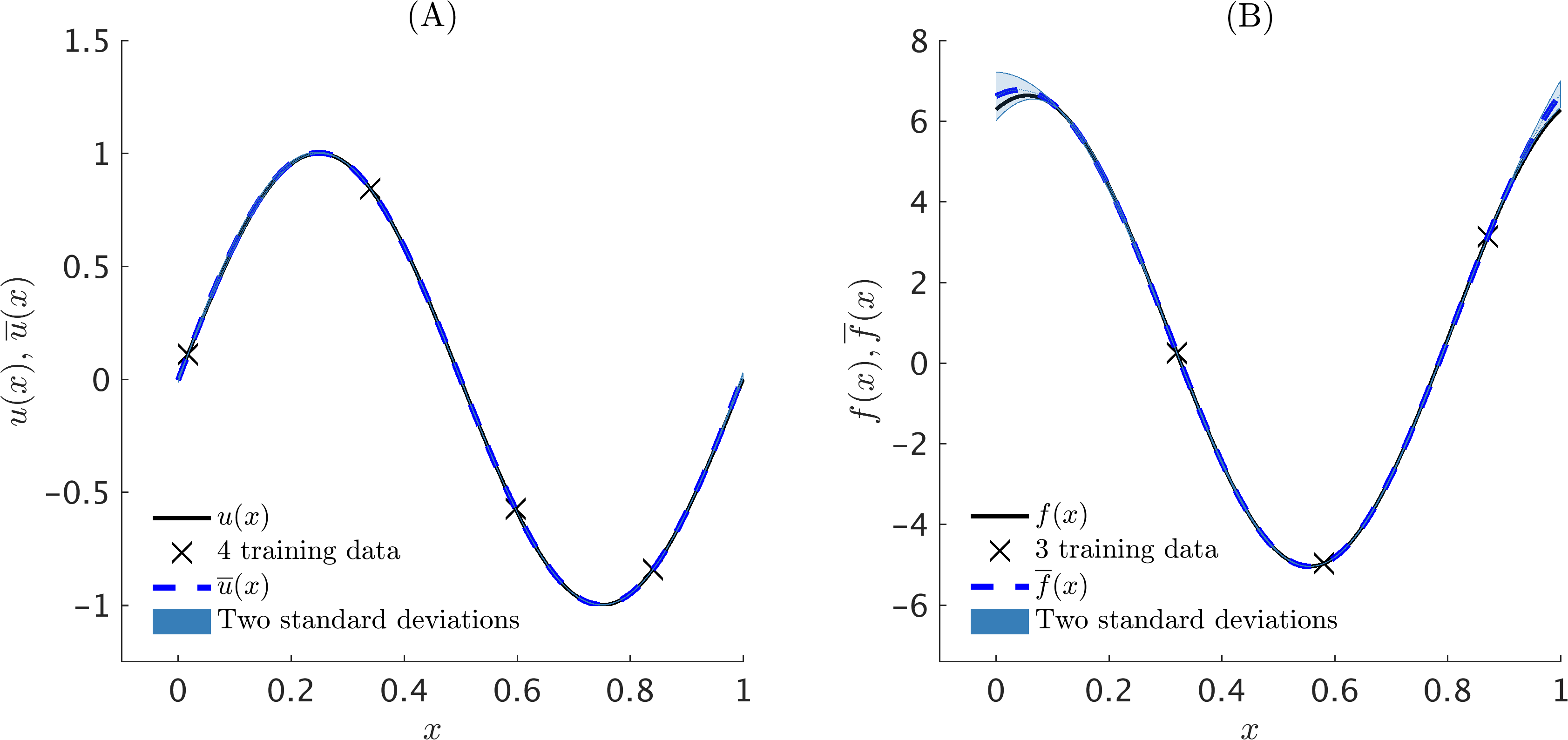}

\vspace{0.5cm}
\includegraphics[width=0.9\textwidth]{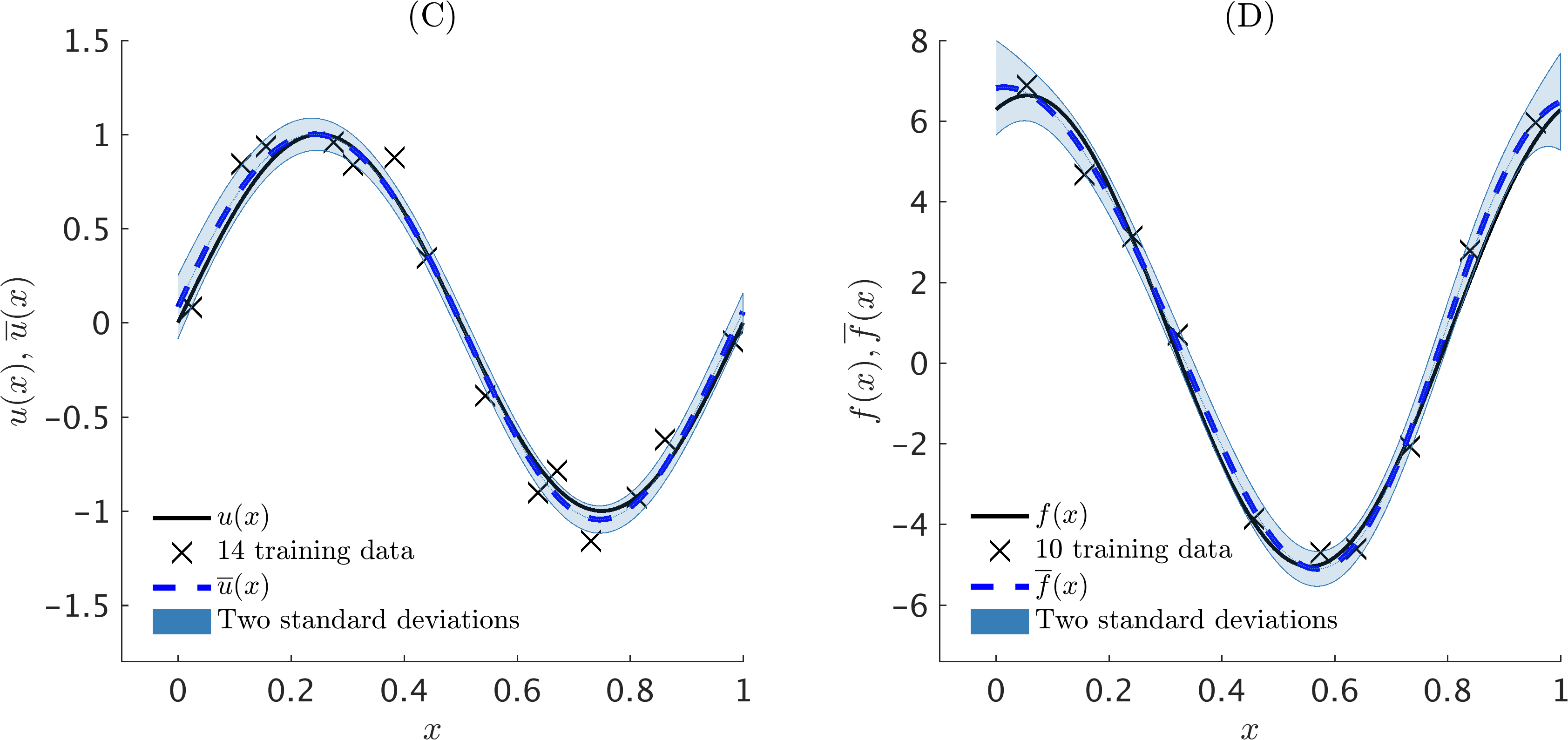}
\caption{{\bf Integro-differential equation in 1D:} {\bf (A)} Exact left-hand-side function $u(x)$, ``noise-free" training data $\{\bm{x}_u,\bm{y}_u\}$, predictive mean $\overline{u}(x)$, and two-standard-deviation band around the mean. {\bf (B)} Exact right-hand-side function $f(x)$, ``noise-free" training data $\{\bm{x}_f,\bm{y}_f\}$, predictive mean $\overline{f}(x)$, and two-standard-deviation band around the mean. {\bf (C)} Exact left-hand-side function $u(x)$, ``noisy" training data $\{\bm{x}_u,\bm{y}_u\}$, predictive mean $\overline{u}(x)$, and two-standard-deviation band around the mean. {\bf (D)} Exact right-hand-side function $f(x)$, ``noisy" training data $\{\bm{x}_f,\bm{y}_f\}$, predictive mean $\overline{f}(x)$, and two-standard-deviation band around the mean.} \label{fig:integro_differential}
\end{figure}

\subsection{Integro-differential equation in 1D} Consider the following integro-differential equation,
\begin{equation}\label{eq:Integro}
\mathcal{L}^{(\alpha,\beta)}_x u(x) := \frac{d}{d x} u(x) + \alpha u(x) + \beta \int_0^x u(\xi)d\xi = f(x).
\end{equation}
Note that for $(\alpha,\beta) = (2,5)$, the functions $u(x) = \sin(2 \pi x)$ and $f(x) = 2 \pi \cos(2 \pi x) + \frac{5}{\pi}\sin(\pi x)^2 + 2\sin(2\pi x)$ satisfy this equation. In the following, the parameters $(\alpha,\beta)$ will be infered from two types of data, namely, noise-free and noisy observations.

\subsubsection*{Noise-free data}
Assume that the noise-free data $\{\bm{x}_u, \bm{y}_u\}$, $\{\bm{x}_f, \bm{y}_f\}$ on $u(x)$, $f(x)$ are generated according to $\bm{y}_u = u(\bm{x}_u)$, $\bm{y}_f = f(\bm{x}_f)$ with $\bm{x}_u$, $\bm{x}_f$ constituting of $n_u = 4$, $n_f = 3$ data points chosen at random in the interval $[0,1]$, respectively. Given these noise-free training data, the algorithm learns the parameters $(\alpha,\beta)$ to have values $(2.012627,\ 4.977879)$. It should be emphasized that the algorithm is able to learn the parameters of the operator using only $7$ training data. Moreover, the resulting posterior distributions for $u(x)$ and $f(x)$ are depicted in Figure \ref{fig:integro_differential}(A, B). The posterior variances could be used as good indicators of how uncertain one should be about the estimated parameters and predictions made based on these parameters.

\subsubsection*{Noisy data}
Consider the case where the noisy data $\{\bm{x}_u, \bm{y}_u\}$, $\{\bm{x}_f, \bm{y}_f\}$ on $u(x)$, $f(x)$ are generated according to $\bm{y}_u = u(\bm{x}_u) + \bm{\epsilon}_u$, $\bm{y}_f = f(\bm{x}_f) + \bm{\epsilon}_f$ with $\bm{x}_u$, $\bm{x}_f$ constituting of $n_u = 14$, $n_f = 10$ data points chosen at random in the interval $[0,1]$, respectively. Here, the noise $\bm{\epsilon}_u$ and $\bm{\epsilon}_f$ are randomly generated according to the normal distributions $\mathcal{N}(\bm{0},0.1^2\ \bm{I}_{n_u})$ and $\mathcal{N}(\bm{0},0.5^2\ \bm{I}_{n_f})$, respectively. Given these noisy training data, the algorithm learns the parameters $(\alpha,\beta)$ to have values $(2.073054,\ 5.627249)$. It should be emphasized that for this example the data is deliberately chosen to have a sizable noise. This highlights the ability of the method to handle highly noisy observations without any modifications. The resulting posterior distributions for $u(x)$ and $f(x)$ are depicted in Figure \ref{fig:integro_differential}(C, D). The posterior variances not only quantify scarcity in observations but also signify noise levels in data.

\begin{figure}
\centering
\includegraphics[width=0.825\textwidth]{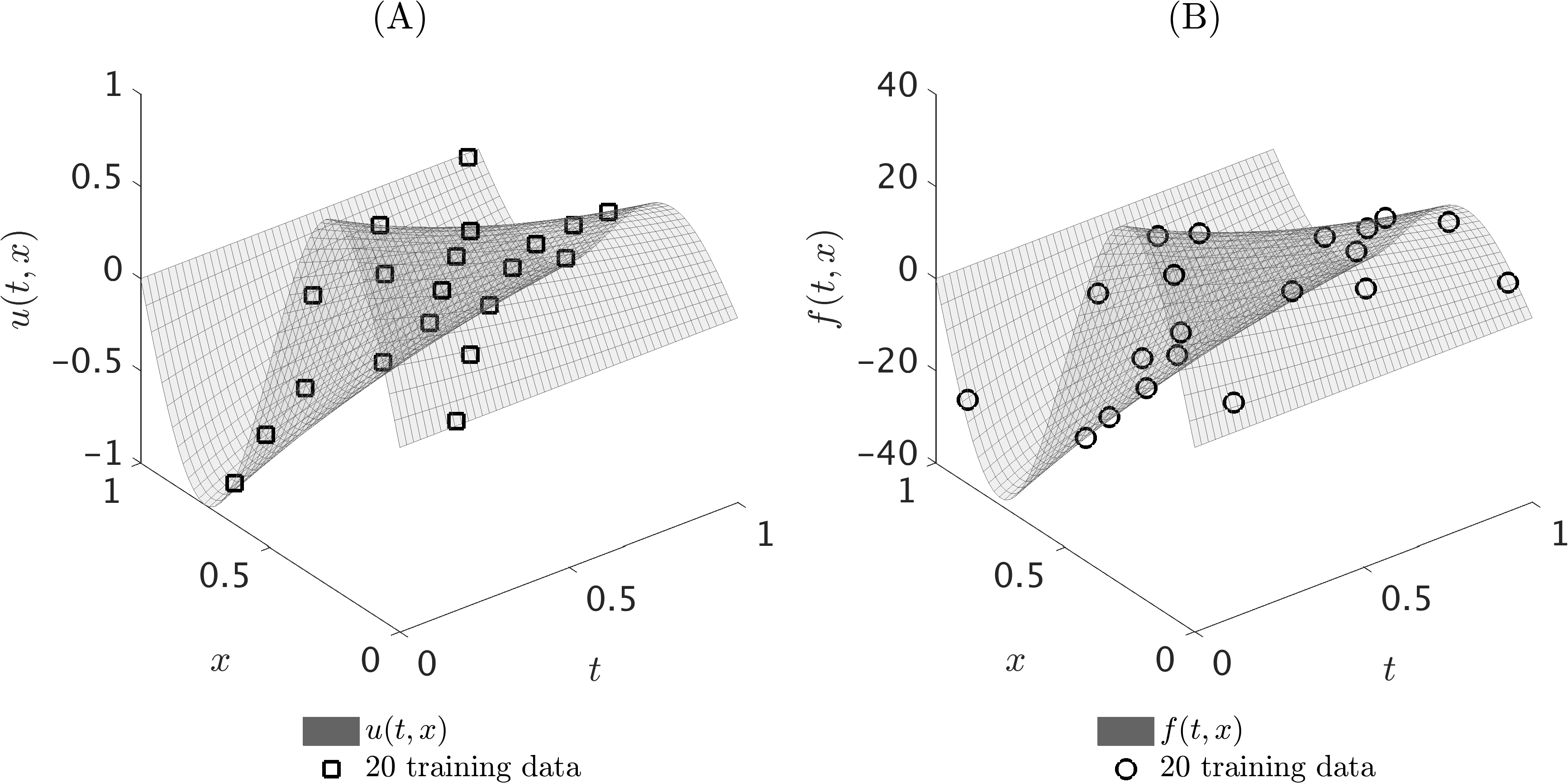}

\vspace{0.5cm}
\includegraphics[width=0.825\textwidth]{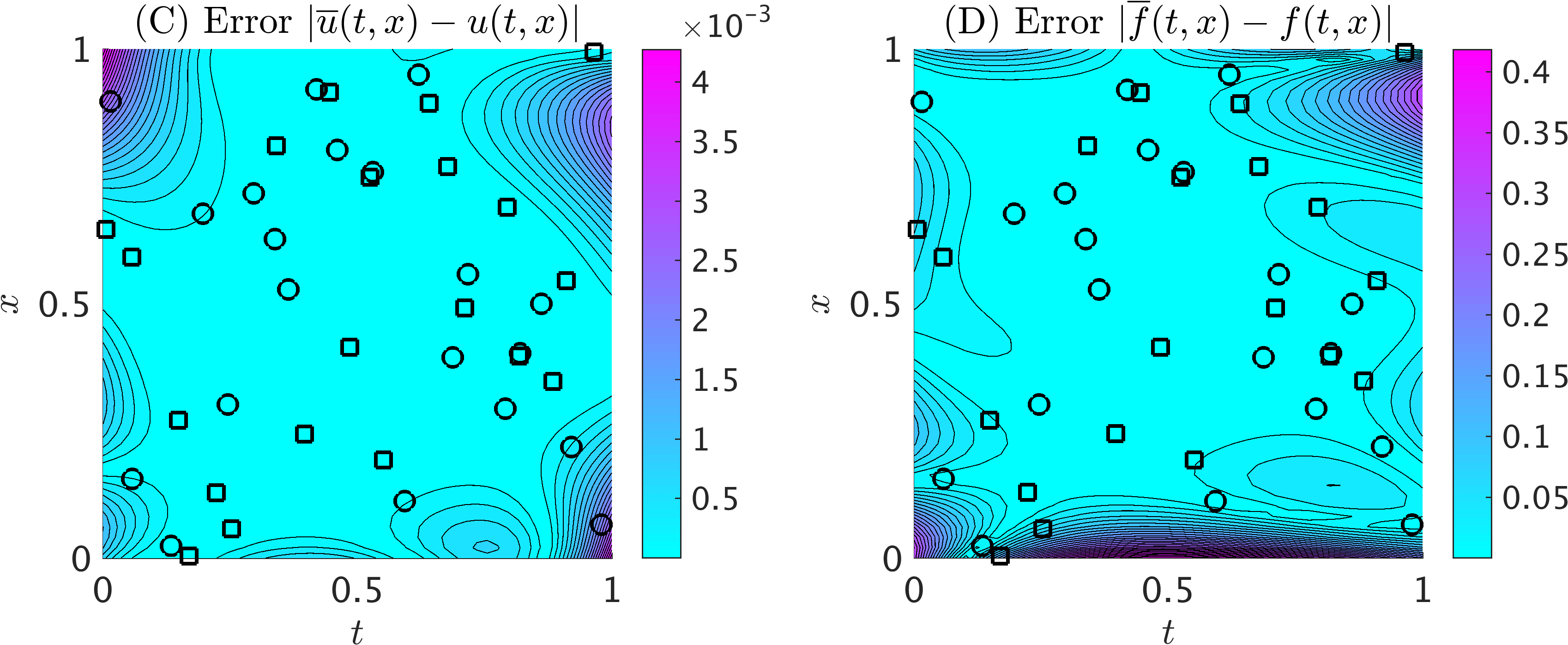}

\vspace{0.5cm}
\includegraphics[width=0.825\textwidth]{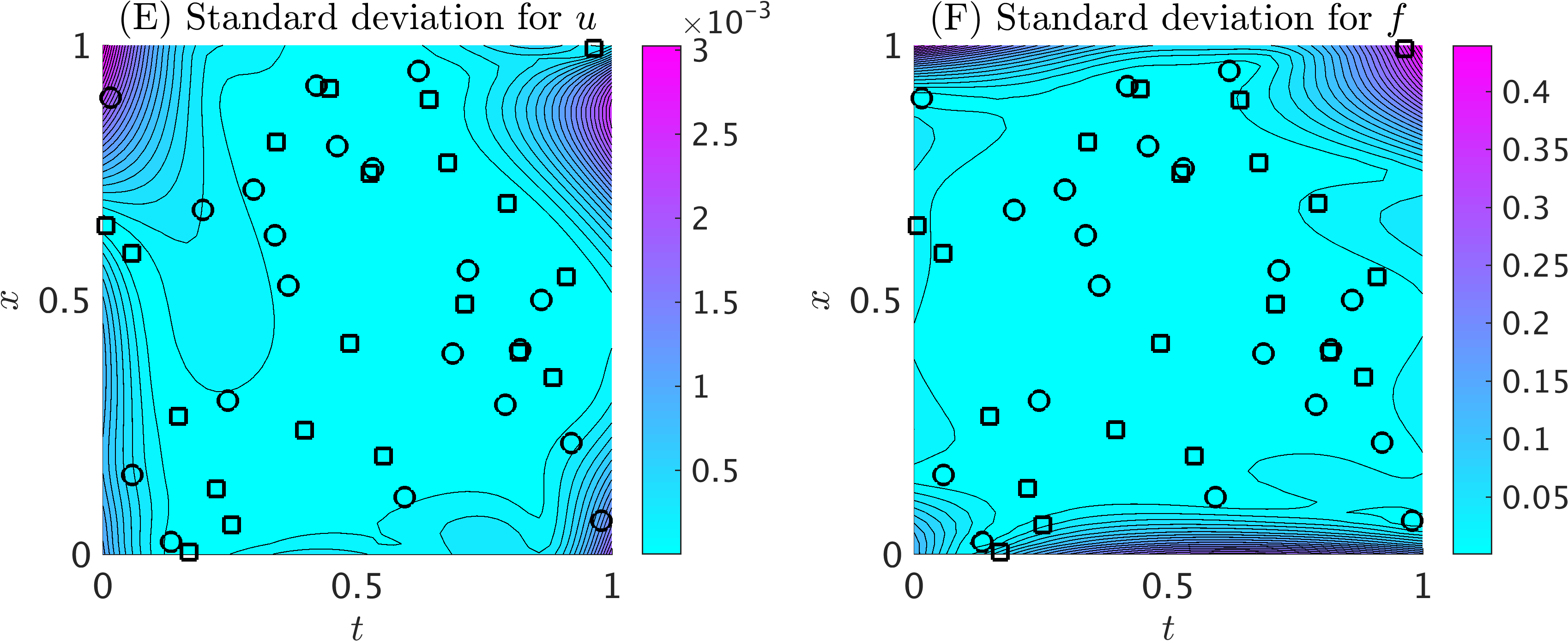}

\caption{{\bf Heat equation:} {\bf (A)} Exact left-hand-side function $u(t,x)$ and training data $\{(\bm{t}_u,\bm{x}_u),\bm{y}_u\}$. {\bf (B)} Exact right-hand-side function $f(t,x)$ and training data $\{(\bm{t}_f,\bm{x}_f),\bm{y}_f\}$. {\bf (C)} Absolute point-wise error betrween the predictive mean $\overline{u}(t,x)$ and the exact function $u(t,x)$. The relative $L_2$ error for the left-hand-side function is $1.250278\mathrm{e}{-03}$. {\bf (D)} Absolute point-wise error betrween the predictive mean $\overline{f}(t,x)$ and the exact function $f(t,x)$. The relative $L_2$ error for the right-hand-side function is $4.167404\mathrm{e}{-03}$. {\bf (E)}, {\bf (F)} Standard deviations $s_u(t,x)$ and $s_f(t,x)$ for $u$ and $f$, respectively.} \label{fig:heat}
\end{figure}

\subsection{Heat Equation} This example is chosen to highlight the capability of the proposed
framework to handle time-dependent problems using only scattered space-time observations. To this end, consider the heat equation
\[
\mathcal{L}_{(t,x)}^\alpha u(t,x) := \frac{\partial}{\partial t} u(t,x) - \alpha \frac{\partial^2}{\partial x^2} u(t,x) = f(t,x).
\]
Note that for $\alpha = 1$, the functions $f(t,x) = e^{-t}(4\pi^2-1)\sin(2\pi x)$ and $u(t,x) = e^{-t}\sin(2 \pi x)$ satisfy this equation. Assume that the noise-free data $\{(\bm{t}_u,\bm{x}_u), \bm{y}_u\}$, $\{(\bm{t}_f,\bm{x}_f), \bm{y}_f\}$ on $u(t,x)$, $f(t,x)$ are generated according to $\bm{y}_u = u(\bm{t}_u,\bm{x}_u)$, $\bm{y}_f = f(\bm{t}_f,\bm{x}_f)$ with $(\bm{t}_u,\bm{x}_u)$, $(\bm{t}_f,\bm{x}_f)$ constituting of $n_u = n_f = 20$ data points chosen at random in the domain $[0,1]^2$, respectively. Given these training data, the algorithm learns the parameter $\alpha$ to have value $0.999943$. The resulting posterior distributions for $u(t,x)$ and $f(t,x)$ are depicted in Figure \ref{fig:heat}. A visual inspection of this figure illustrates how closely uncertainty in predictions measured by posterior variances (see Figure \ref{fig:heat}(E, F)) correlate with prediction errors (see Figure \ref{fig:heat}(C, D)). Remarkably, the proposed methodology circumvents the need for temporal discretization, and is essentially immune to any restrictions arising due to time-stepping, e.g., the fundamental consistency and stability issues in classical numerical analysis.


\begin{figure}
\centering
\includegraphics[width=0.9\textwidth]{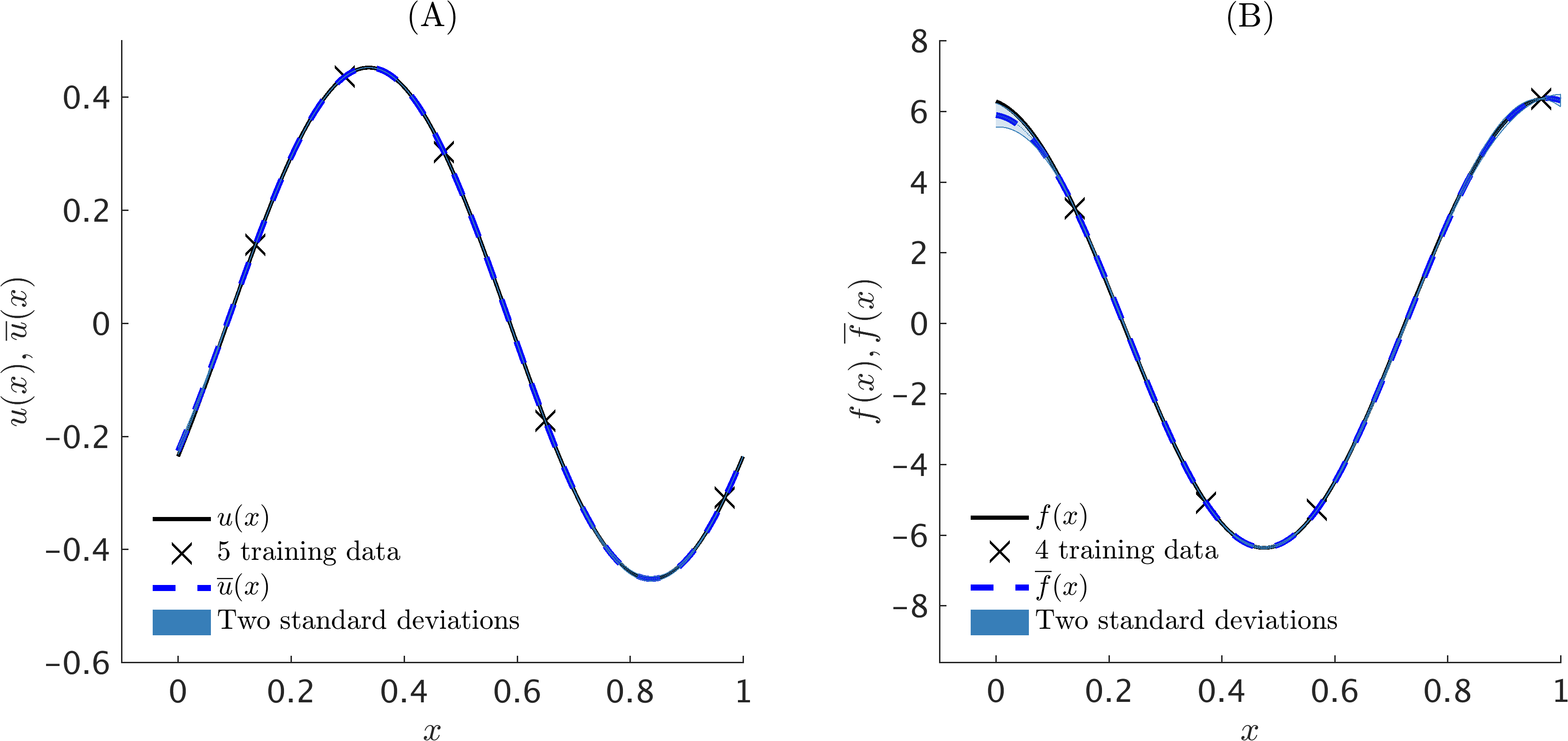}
\caption{{\bf Fractional equation in 1D:} {\bf (A)} Exact left-hand-side function $u(x)$, training data $\{\bm{x}_u,\bm{y}_u\}$, predictive mean $\overline{u}(x)$, and two-standard-deviation band around the mean. {\bf (B)} Exact right-hand-side function $f(x)$, training data $\{\bm{x}_f,\bm{y}_f\}$, predictive mean $\overline{f}(x)$, and two-standard-deviation band around the mean.} \label{fig:fractional}
\end{figure}

\subsection{Fractional Equation} Consider the one dimensional fractional equation
\[
\mathcal{L}^{\alpha}_x u(x) = {}_{-\infty}D_x^\alpha u(x) - u(x) = f(x),
\]
where $\alpha\in\mathbb{R}$ is the fractional order of the operator that is defined in the Riemann-Liouville sense \cite{podlubny1998fractional}. Fractional operators often arise in modeling anomalous diffusion processes. Their non-local behavior poses serious computational challenges as it involves costly convolution operations for resolving the underlying non-Markovian dynamics \cite{podlubny1998fractional}. However, the machine leaning approach pursued in this work bypasses the need for numerical discretization, hence, overcomes these computational bottlenecks, and can seamlessly handle all such linear cases without any modifications. The only technicality induced by fractional operators has to do with deriving the kernel $k_{ff}(x,x';\theta,\alpha)$ in Eq.~\ref{eq:kernelk}. Here, $k_{ff}(x,x';\theta,\alpha)$ was obtained by taking the inverse Fourier transform  \cite{podlubny1998fractional} of 
\[
[(-iw)^{\alpha}(-iw')^{\alpha} - (-iw)^{\alpha} - (-iw')^{\alpha} + 1]\widehat{k}_{uu}(w,w';\theta), 
\]
where $\widehat{k}_{uu}(w,w';\theta)$ is the Fourier transform of the kernel $k_{uu}(x,x';\theta)$. Similarly, one can obtain $k_{uf}(x,x';\theta,\alpha)$ and $k_{fu}(x,x';\theta,\alpha)$.\\

Note that for $\alpha = \sqrt{2}$, the functions $u(x) = \frac{1}{2} e^{-2 i \pi  x} \left(\frac{(2 \pi +i) e^{4 i \pi  x}}{-1+(2 i \pi )^{\sqrt{2}}}+\frac{2
   \pi -i}{-1+(-2 i \pi )^{\sqrt{2}}}\right)$ and $f(x) = 2 \pi  \cos (2 \pi  x)-\sin (2 \pi  x)$ satisfy the fractional equation. Assume that the noise-free data $\{\bm{x}_u, \bm{y}_u\}$, $\{\bm{x}_f, \bm{y}_f\}$ on $u(x)$, $f(x)$ are generated according to $\bm{y}_u = u(\bm{x}_u)$, $\bm{y}_f = f(\bm{x}_f)$ with $\bm{x}_u$, $\bm{x}_f$ constituting of $n_u = 5$, $n_f = 4$ data points chosen at random in the interval $[0,1]$, respectively. Given these noise-free training data, the algorithm learns the parameter $\alpha$ to have value $1.412104$. The resulting posterior distributions for $u(x)$ and $f(x)$ are depicted in Figure \ref{fig:fractional}.

\begin{figure}[t]
\centering
\includegraphics[width=\textwidth]{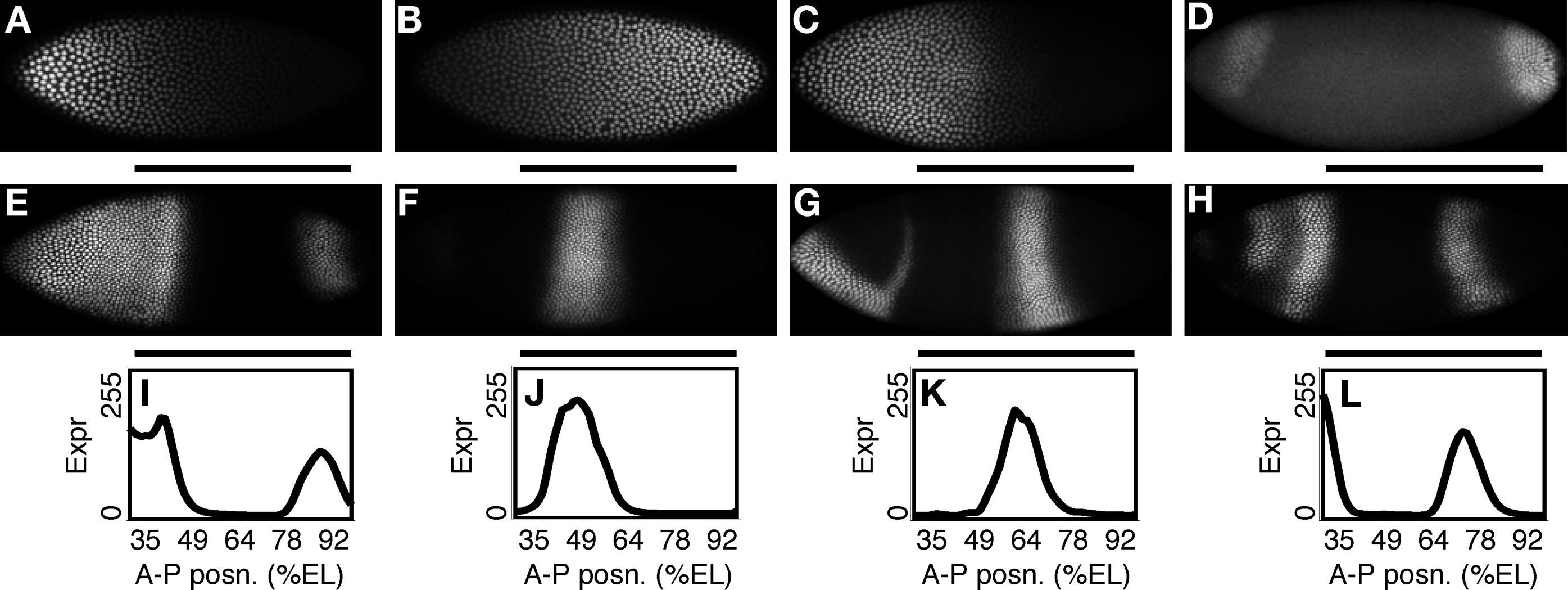}
\caption{{\bf Maternal and Gap Gene Expression (see \cite{perkins2006reverse})}: 
{\bf (A--C)} Drosophila embryos at early blastoderm stage (cleavage cycle 13) fluorescently stained for Bcd {\bf(A)}, Cad {\bf (B)}, and Hb {\bf (C)} protein.
{\bf (D--H)} Drosophila embryos at late blastoderm stage (late cleavage cycle 14A) fluorescently stained for Tll {\bf (D)}, Hb {\bf (E)}, Kr {\bf (F)}, Kni {\bf (G)}, and Gt {\bf (H)} protein. Anterior is to the left, dorsal is up. Black bars indicate the modeled A-P extent.
{\bf (I--L)} Mean relative gap protein concentration as a function of A-P position (measured in percent embryo length) for Hb {\bf (I)}, Kr {\bf (J)}, Kni {\bf (K)}, and Gt {\bf (L)}. Expression levels are from images and are unitless, ranging from 0 to 255. Images and expression profiles are from the FlyEx database \cite{poustelnikova2004database}. {\bf Embryo IDs:} bd3 (A,B), hz30 (C), tb6 (D), kf9 (E), kd17 (F), fq1 (G), nk5 (H).
{\bf Abbreviations:} A-P, anterior-posterior; Bcd, Bicoid; Cad, Caudal; Gt, Giant; Hb, hunchback; Kni, Knirps; Kr, Kr{\"u}ppel; Tll, tailless.
} \label{fig:GapGene}
\end{figure}

\subsection{Drosophila melanogaster gap gene dynamics \cite{perkins2006reverse, alvarez2013linear}} The gap gene dynamics of protein $a \in \{Hb, Kr, Gt, Kni\}$ (see figure \ref{fig:GapGene}) can be modeled using a reaction-diffusion partial differential equation
\[
\mathcal{L}_{(t,x)}^{(\lambda^a,D^a)} u^a(t,x) = \frac{\partial}{\partial t}u^a(t,x) + \lambda^a u^a(t,x) - D^a \frac{\partial^2}{\partial x^2}u^a(t,x) = f^a(t,x),
\]
where $u^a(t,x)$ denotes the relative concentration of gap protein $a$ (unitless, ranging from 0 to 255) at space point $x$ (from 35\% to 92\% of embryo length) and time $t$ (0 min to 68 min after the start of cleavage cycle 13). Here, $\lambda^a$ and $D^a$ are decay and diffusion rates of protein $a$, respectively. Moreover, the right-hand-side is given by
\[
f^a(t,x) := \zeta(t)P^a(t,x),
\]
where the term
\[
\zeta(t) = \left\{\begin{array}{cl}
0.5 & ~0 \text{ min } \leq t < 16 \text{ min } \\ 
0.0 & 16 \text{ min } \leq t < 21 \text{ min } \\ 
1.0 & 21 \text{ min } \leq t
\end{array} \right.
\]
models the doubling of nuclei and shutdown of transcription during mitosis and 
\[
P^a(t,x) = R^a g\left(\sum_b T^{ab} u^b(t,x) + h^a\right)
\]
specifies the production rate of protein $a$. The model combines the processes of transcription and translation into a single production process $P^a(t,x)$. Here, $R^a$ is the maximum production rate, 
\[
g(u) = \frac{1}{2}\left(\frac{u}{\sqrt{u^2+1}}+1\right),
\]
and $b \in \{Bcd, Cad, Hb, Kr, Gt, Kni, Tll\}$ ranges over the seven genes (see figure \ref{fig:GapGene}). The regulatory weights $T^{ab}$, encode the effect protein $b$ has on the production rate of protein $a$. If $T^{ab} > 0$ (or $T^{ab} < 0$), then gene $b$ is interpreted as being an activator (or a repressor) of gene $a$.\\

\begin{table}[h]
\centering
\label{table}
\resizebox{\textwidth}{!}{%
\begin{tabular}{|c|c|ccccccc|c|}
\hline
 & Max prod. & \multicolumn{7}{c|}{Regulatory weights ($T^{ab}$)} & Bias \\
Gene & rate ($R^a$) & Bcd  & Cad  & Hb  & Kr & Gt & Kni & Tll & ($h^a$) \\ \hline
Hb & 32.03 & 0.1114  & -0.0054  & 0.0293  & -0.0124 & 0.0553 & -0.3903 & 0.0144 & -3.5 \\
Kr & 16.70 & 0.1173  & 0.0215  & -0.0498  & 0.0755 & -0.0141 & -0.0666 & -1.2036 & -3.5 \\
Gt & 25.15 & 0.0738  & 0.0180  & -0.0008  & -0.0758 & 0.0157 & 0.0056 & -0.0031 & -3.5 \\
Kni & 16.12 & 0.2146  & 0.0210  & -0.1891  & -0.0458 & -0.1458 & 0.0887 & -0.3028 & -3.5 \\ \hline
\end{tabular}
}
\caption{Parameters $R^a$, $T^{ab}$, and $h^a$ are assumed to be exogenously given and their values are taken from \cite{perkins2006reverse}.}
\end{table}

\begin{figure}
\centering
\includegraphics[width=\textwidth]{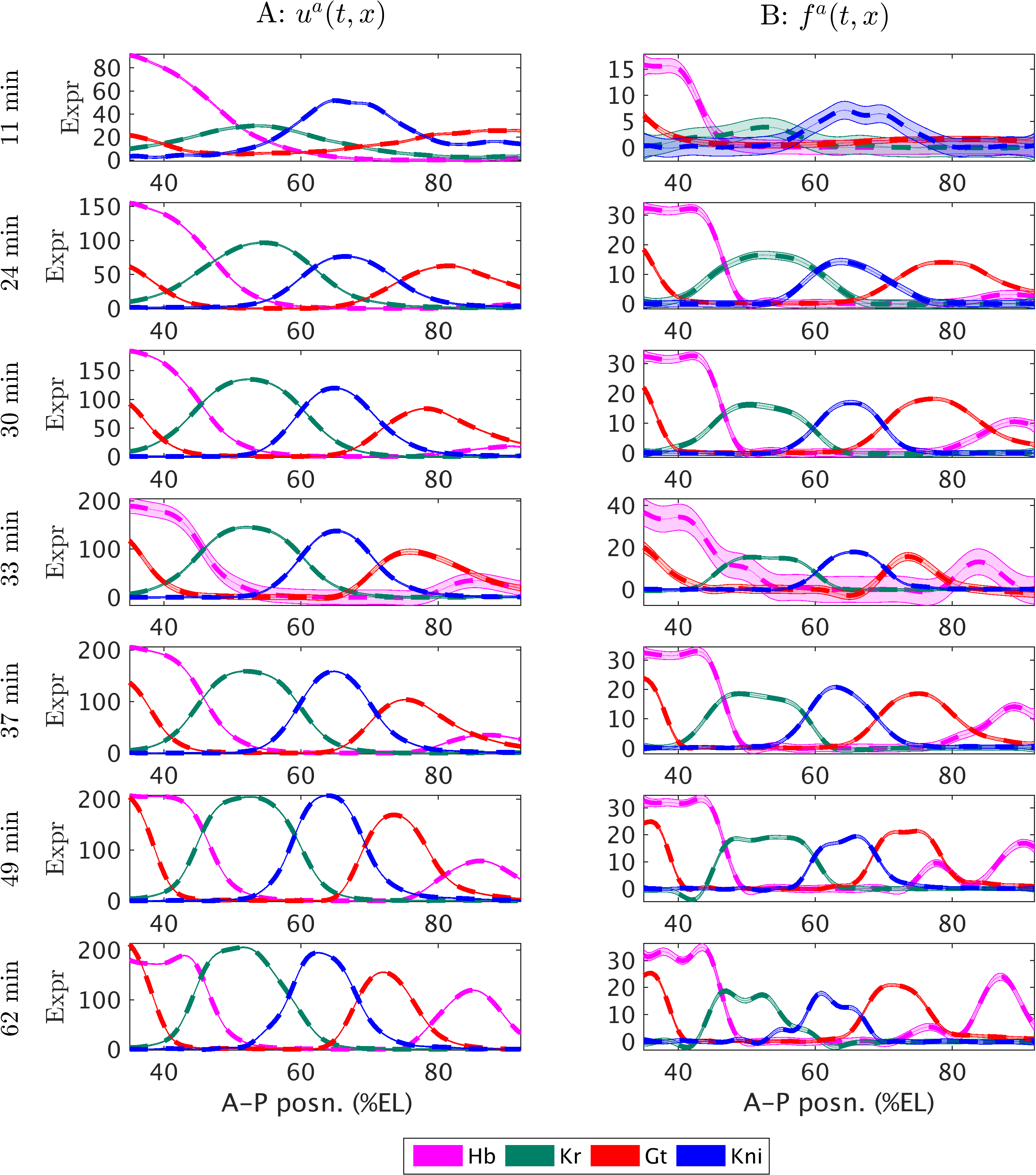}
\caption{{\bf Predictive expression at seven points in time}: 
{\bf (A)} Predictive mean expression along with the two-standard-deviations band around the mean for Hb, Kr, Gt, and Kni. The vertical axis represents relative protein concentration corresponding to fluorescence intensity from quantitative gene expression data \cite{poustelnikova2004database,perkins2006reverse}. {\bf (B)} Predictive mean along with the two-standard-deviations band around the mean for the right-hand-side function corresponding to Hb, Kr, Gt, and Kni. The horizontal axis in each plot is A-P position, ranging from 35\% to 92\% of embryo length. No data points are available at time $t=33$ min.
} \label{fig:GapGene_solution}
\end{figure}

This work assumes the maximum production rate $R^a$, the regulatory weights $T^{ab}$, and the bias or offset $h^a$ to be specified as in table \ref{table} and seeks to learn the decay $\lambda^a$ and diffusion $D^a$ rates of protein $a$. In fact, table \ref{table_results} summarizes the values learned by the algorithm for these parameters and figure \ref{fig:GapGene_solution} depicts the corresponding posterior distributions for $u^a(t,x)$ and $f^a(t,x)$. Indeed, figure \ref{fig:GapGene_solution} gives a good indication of how certain one could be about the estimated parameters and the predictions made based on them.\\

\begin{table}[h]
\centering
\label{table_results}
\begin{tabular}{|c|c|c|}
\hline
 & Decay &  Diff.\\
Gene & ($\lambda^a$) & ($D^a$) \\ \hline
Hb & 0.1606  &  0.3669   \\
Kr & 0.0797  &  0.4490   \\
Gt & 0.1084  &  0.4543   \\
Kni & 0.0807  &   0.2683  \\ \hline
\end{tabular}
\caption{Inferred parameter values for the decay $\lambda^a$ and diffusion $D^a$ rates of protein $a$.}
\end{table}

\section{Discussion} In summary, this work introduced a possibly disruptive probabilistic technology for learning general parametric linear equations from noisy data. This generality was demostrated using various bechmark problems with utterly different attributes along with an example application in functional genomics. Furthermore, the current methodology can be applied to inverse problems involving characterization of materials, tomography and electrophysiology, design of effective metamaterials, etc. The methodology can be straightforwardly generalized to address data with multiple levels of fidelity \cite{kennedy2000predicting, le2013multi} and equations with variable coefficients and complex geometries. Non-Gaussian and input-dependent noise models (e.g., student-t, heteroscedastic, etc.) \cite{williams2006gaussian} can also be accommodated. Moreover, systems of linear integro-differential equations can be addressed using multi-output Gaussian process regressions \cite{osborne2008towards, alvarez2009sparse, boyle2004dependent}. These scenarios are all feasible because they do not affect the key observation that any linear transformation of a Gaussian process is still a Gaussian process. In its current form, despite its generality regarding linear equations, the proposed framework cannot deal with non-linear equations. However, some specific non-linear operators can be addressed with extensions of the current framework by transforming such equations into systems of linear equations \cite{zwanzig1960ensemble, chorin2000optimal} -- albeit in high dimensions. In the end, the proposed methodology in this work, being essentially a regression technology, is suitable for resolving such high-dimensional problems.





\bibliographystyle{model1-num-names}
\bibliography{mybib.bib}

\begin{thebibliography}{42}
\expandafter\ifx\csname natexlab\endcsname\relax\def\natexlab#1{#1}\fi
\providecommand{\bibinfo}[2]{#2}
\ifx\xfnm\relax \def\xfnm[#1]{\unskip,\space#1}\fi
\bibitem[{Kaipio and Somersalo(2006)}]{kaipio2006statistical}
\bibinfo{author}{J.~Kaipio}, \bibinfo{author}{E.~Somersalo},
  \bibinfo{title}{Statistical and computational inverse problems}, volume
  \bibinfo{volume}{160}, \bibinfo{publisher}{Springer Science \& Business
  Media}, \bibinfo{year}{2006}.
\bibitem[{Stuart(2010)}]{stuart2010inverse}
\bibinfo{author}{A.~M. Stuart},
\newblock \bibinfo{title}{Inverse problems: a bayesian perspective},
\newblock \bibinfo{journal}{Acta Numerica} \bibinfo{volume}{19}
  (\bibinfo{year}{2010}) \bibinfo{pages}{451--559}.
\bibitem[{Williams and Rasmussen(2006)}]{williams2006gaussian}
\bibinfo{author}{C.~K. Williams}, \bibinfo{author}{C.~E. Rasmussen},
\newblock \bibinfo{title}{Gaussian processes for machine learning},
\newblock \bibinfo{journal}{the MIT Press} \bibinfo{volume}{2}
  (\bibinfo{year}{2006}) \bibinfo{pages}{4}.
\bibitem[{Murphy(2012)}]{murphy2012machine}
\bibinfo{author}{K.~P. Murphy}, \bibinfo{title}{Machine learning: a
  probabilistic perspective}, \bibinfo{publisher}{MIT press},
  \bibinfo{year}{2012}.
\bibitem[{Diaconis(1988)}]{diaconis1988bayesian}
\bibinfo{author}{P.~Diaconis},
\newblock \bibinfo{title}{Bayesian numerical analysis},
\newblock \bibinfo{journal}{Statistical decision theory and related topics IV}
  \bibinfo{volume}{1} (\bibinfo{year}{1988}) \bibinfo{pages}{163--175}.
\bibitem[{Vapnik(2013)}]{vapnik2013nature}
\bibinfo{author}{V.~Vapnik}, \bibinfo{title}{The nature of statistical learning
  theory}, \bibinfo{publisher}{Springer Science \& Business Media},
  \bibinfo{year}{2013}.
\bibitem[{Sch{\"o}lkopf and Smola(2002)}]{scholkopf2002learning}
\bibinfo{author}{B.~Sch{\"o}lkopf}, \bibinfo{author}{A.~J. Smola},
  \bibinfo{title}{Learning with kernels: support vector machines,
  regularization, optimization, and beyond}, \bibinfo{publisher}{MIT press},
  \bibinfo{year}{2002}.
\bibitem[{Tipping(2001)}]{tipping2001sparse}
\bibinfo{author}{M.~E. Tipping},
\newblock \bibinfo{title}{Sparse bayesian learning and the relevance vector
  machine},
\newblock \bibinfo{journal}{The journal of machine learning research}
  \bibinfo{volume}{1} (\bibinfo{year}{2001}) \bibinfo{pages}{211--244}.
\bibitem[{Tikhonov(1963)}]{tikhonov1963solution}
\bibinfo{author}{A.~Tikhonov},
\newblock \bibinfo{title}{Solution of incorrectly formulated problems and the
  regularization method},
\newblock in: \bibinfo{booktitle}{Soviet Math. Dokl.},
  volume~\bibinfo{volume}{5}, pp. \bibinfo{pages}{1035--1038}.
\bibitem[{Tikhonov and Arsenin(1977)}]{Tikhonov/Arsenin/77}
\bibinfo{author}{A.~N. Tikhonov}, \bibinfo{author}{V.~Y. Arsenin},
  \bibinfo{title}{Solutions of Ill-posed problems},
  \bibinfo{publisher}{W.H.~Winston}, \bibinfo{year}{1977}.
\bibitem[{Poggio and Girosi(1990)}]{poggio1990networks}
\bibinfo{author}{T.~Poggio}, \bibinfo{author}{F.~Girosi},
\newblock \bibinfo{title}{Networks for approximation and learning},
\newblock \bibinfo{journal}{Proceedings of the IEEE} \bibinfo{volume}{78}
  (\bibinfo{year}{1990}) \bibinfo{pages}{1481--1497}.
\bibitem[{Franke and Schaback(1998)}]{franke1998solving}
\bibinfo{author}{C.~Franke}, \bibinfo{author}{R.~Schaback},
\newblock \bibinfo{title}{Solving partial differential equations by collocation
  using radial basis functions},
\newblock \bibinfo{journal}{Applied Mathematics and Computation}
  \bibinfo{volume}{93} (\bibinfo{year}{1998}) \bibinfo{pages}{73--82}.
\bibitem[{Fasshauer and Ye(2013)}]{fasshauer2013kernel}
\bibinfo{author}{G.~E. Fasshauer}, \bibinfo{author}{Q.~Ye},
\newblock \bibinfo{title}{A kernel-based collocation method for elliptic
  partial differential equations with random coefficients},
\newblock in: \bibinfo{booktitle}{Monte Carlo and Quasi-Monte Carlo Methods
  2012}, \bibinfo{publisher}{Springer}, \bibinfo{year}{2013}, pp.
  \bibinfo{pages}{331--347}.
\bibitem[{Owhadi(2015)}]{owhadi2015bayesian}
\bibinfo{author}{H.~Owhadi},
\newblock \bibinfo{title}{Bayesian numerical homogenization},
\newblock \bibinfo{journal}{Multiscale Modeling \& Simulation}
  \bibinfo{volume}{13} (\bibinfo{year}{2015}) \bibinfo{pages}{812--828}.
\bibitem[{Cockayne et~al.(2016)Cockayne, Oates, Sullivan, and
  Girolami}]{cockayne2016probabilistic}
\bibinfo{author}{J.~Cockayne}, \bibinfo{author}{C.~Oates},
  \bibinfo{author}{T.~Sullivan}, \bibinfo{author}{M.~Girolami},
\newblock \bibinfo{title}{Probabilistic meshless methods for partial
  differential equations and bayesian inverse problems},
\newblock \bibinfo{journal}{arXiv preprint arXiv:1605.07811}
  (\bibinfo{year}{2016}).
\bibitem[{{\'A}lvarez et~al.(2013){\'A}lvarez, Luengo, and
  Lawrence}]{alvarez2013linear}
\bibinfo{author}{M.~A. {\'A}lvarez}, \bibinfo{author}{D.~Luengo},
  \bibinfo{author}{N.~D. Lawrence},
\newblock \bibinfo{title}{Linear latent force models using gaussian processes},
\newblock \bibinfo{journal}{IEEE transactions on pattern analysis and machine
  intelligence} \bibinfo{volume}{35} (\bibinfo{year}{2013})
  \bibinfo{pages}{2693--2705}.
\bibitem[{Alvarez et~al.(2009)Alvarez, Luengo, and
  Lawrence}]{alvarez2009latent}
\bibinfo{author}{M.~A. Alvarez}, \bibinfo{author}{D.~Luengo},
  \bibinfo{author}{N.~D. Lawrence},
\newblock \bibinfo{title}{Latent force models.},
\newblock in: \bibinfo{booktitle}{AISTATS}, volume~\bibinfo{volume}{12}, pp.
  \bibinfo{pages}{9--16}.
\bibitem[{Lawrence(2005)}]{lawrence2005probabilistic}
\bibinfo{author}{N.~Lawrence},
\newblock \bibinfo{title}{Probabilistic non-linear principal component analysis
  with gaussian process latent variable models},
\newblock \bibinfo{journal}{Journal of Machine Learning Research}
  \bibinfo{volume}{6} (\bibinfo{year}{2005}) \bibinfo{pages}{1783--1816}.
\bibitem[{Lawrence(2004)}]{lawrence2004gaussian}
\bibinfo{author}{N.~D. Lawrence},
\newblock \bibinfo{title}{Gaussian process latent variable models for
  visualisation of high dimensional data},
\newblock \bibinfo{journal}{Advances in neural information processing systems}
  \bibinfo{volume}{16} (\bibinfo{year}{2004}) \bibinfo{pages}{329--336}.
\bibitem[{Titsias and Lawrence(2010)}]{titsias2010bayesian}
\bibinfo{author}{M.~K. Titsias}, \bibinfo{author}{N.~D. Lawrence},
\newblock \bibinfo{title}{Bayesian gaussian process latent variable model.},
\newblock in: \bibinfo{booktitle}{AISTATS}, pp. \bibinfo{pages}{844--851}.
\bibitem[{Higdon(2002)}]{higdon2002space}
\bibinfo{author}{D.~Higdon},
\newblock \bibinfo{title}{Space and space-time modeling using process
  convolutions},
\newblock in: \bibinfo{booktitle}{Quantitative methods for current
  environmental issues}, \bibinfo{publisher}{Springer}, \bibinfo{year}{2002},
  pp. \bibinfo{pages}{37--56}.
\bibitem[{Boyle and Frean(2004)}]{boyle2004dependent}
\bibinfo{author}{P.~Boyle}, \bibinfo{author}{M.~Frean},
\newblock \bibinfo{title}{Dependent gaussian processes},
\newblock in: \bibinfo{booktitle}{Advances in neural information processing
  systems}, pp. \bibinfo{pages}{217--224}.
\bibitem[{Alvarez and Lawrence(2009)}]{alvarez2009sparse}
\bibinfo{author}{M.~Alvarez}, \bibinfo{author}{N.~D. Lawrence},
\newblock \bibinfo{title}{Sparse convolved gaussian processes for multi-output
  regression},
\newblock in: \bibinfo{booktitle}{Advances in neural information processing
  systems}, pp. \bibinfo{pages}{57--64}.
\bibitem[{Kennedy and O'Hagan(2000)}]{kennedy2000predicting}
\bibinfo{author}{M.~C. Kennedy}, \bibinfo{author}{A.~O'Hagan},
\newblock \bibinfo{title}{Predicting the output from a complex computer code
  when fast approximations are available},
\newblock \bibinfo{journal}{Biometrika} \bibinfo{volume}{87}
  (\bibinfo{year}{2000}) \bibinfo{pages}{1--13}.
\bibitem[{Le~Gratiet and Garnier(2014)}]{le2014recursive}
\bibinfo{author}{L.~Le~Gratiet}, \bibinfo{author}{J.~Garnier},
\newblock \bibinfo{title}{Recursive co-kriging model for design of computer
  experiments with multiple levels of fidelity},
\newblock \bibinfo{journal}{International Journal for Uncertainty
  Quantification} \bibinfo{volume}{4} (\bibinfo{year}{2014}).
\bibitem[{Aronszajn(1950)}]{aronszajn1950theory}
\bibinfo{author}{N.~Aronszajn},
\newblock \bibinfo{title}{Theory of reproducing kernels},
\newblock \bibinfo{journal}{Transactions of the American mathematical society}
  \bibinfo{volume}{68} (\bibinfo{year}{1950}) \bibinfo{pages}{337--404}.
\bibitem[{Saitoh(1988)}]{saitoh1988theory}
\bibinfo{author}{S.~Saitoh}, \bibinfo{title}{Theory of reproducing kernels and
  its applications}, volume \bibinfo{volume}{189},
  \bibinfo{publisher}{Longman}, \bibinfo{year}{1988}.
\bibitem[{Berlinet and Thomas-Agnan(2011)}]{berlinet2011reproducing}
\bibinfo{author}{A.~Berlinet}, \bibinfo{author}{C.~Thomas-Agnan},
  \bibinfo{title}{Reproducing kernel Hilbert spaces in probability and
  statistics}, \bibinfo{publisher}{Springer Science \& Business Media},
  \bibinfo{year}{2011}.
\bibitem[{Liu and Nocedal(1989)}]{liu1989limited}
\bibinfo{author}{D.~C. Liu}, \bibinfo{author}{J.~Nocedal},
\newblock \bibinfo{title}{On the limited memory bfgs method for large scale
  optimization},
\newblock \bibinfo{journal}{Mathematical programming} \bibinfo{volume}{45}
  (\bibinfo{year}{1989}) \bibinfo{pages}{503--528}.
\bibitem[{Rasmussen and Ghahramani(2001)}]{rasmussen2001occam}
\bibinfo{author}{C.~E. Rasmussen}, \bibinfo{author}{Z.~Ghahramani},
\newblock \bibinfo{title}{Occam's razor},
\newblock \bibinfo{journal}{Advances in neural information processing systems}
  (\bibinfo{year}{2001}) \bibinfo{pages}{294--300}.
\bibitem[{Snelson and Ghahramani(2005)}]{snelson2005sparse}
\bibinfo{author}{E.~Snelson}, \bibinfo{author}{Z.~Ghahramani},
\newblock \bibinfo{title}{Sparse gaussian processes using pseudo-inputs},
\newblock in: \bibinfo{booktitle}{Advances in neural information processing
  systems}, pp. \bibinfo{pages}{1257--1264}.
\bibitem[{Hensman et~al.(2013)Hensman, Fusi, and
  Lawrence}]{hensman2013gaussian}
\bibinfo{author}{J.~Hensman}, \bibinfo{author}{N.~Fusi}, \bibinfo{author}{N.~D.
  Lawrence},
\newblock \bibinfo{title}{Gaussian processes for big data},
\newblock \bibinfo{journal}{arXiv preprint arXiv:1309.6835}
  (\bibinfo{year}{2013}).
\bibitem[{Cohn et~al.(1996)Cohn, Ghahramani, and Jordan}]{cohn1996active}
\bibinfo{author}{D.~A. Cohn}, \bibinfo{author}{Z.~Ghahramani},
  \bibinfo{author}{M.~I. Jordan},
\newblock \bibinfo{title}{Active learning with statistical models},
\newblock \bibinfo{journal}{Journal of artificial intelligence research}
  (\bibinfo{year}{1996}).
\bibitem[{Krause and Guestrin(2007)}]{krause2007nonmyopic}
\bibinfo{author}{A.~Krause}, \bibinfo{author}{C.~Guestrin},
\newblock \bibinfo{title}{Nonmyopic active learning of gaussian processes: an
  exploration-exploitation approach},
\newblock in: \bibinfo{booktitle}{Proceedings of the 24th international
  conference on Machine learning}, \bibinfo{organization}{ACM}, pp.
  \bibinfo{pages}{449--456}.
\bibitem[{MacKay(1992)}]{mackay1992information}
\bibinfo{author}{D.~J. MacKay},
\newblock \bibinfo{title}{Information-based objective functions for active data
  selection},
\newblock \bibinfo{journal}{Neural computation} \bibinfo{volume}{4}
  (\bibinfo{year}{1992}) \bibinfo{pages}{590--604}.
\bibitem[{Perkins et~al.(2006)Perkins, Jaeger, Reinitz, and
  Glass}]{perkins2006reverse}
\bibinfo{author}{T.~J. Perkins}, \bibinfo{author}{J.~Jaeger},
  \bibinfo{author}{J.~Reinitz}, \bibinfo{author}{L.~Glass},
\newblock \bibinfo{title}{Reverse engineering the gap gene network of
  drosophila melanogaster},
\newblock \bibinfo{journal}{PLoS Comput Biol} \bibinfo{volume}{2}
  (\bibinfo{year}{2006}) \bibinfo{pages}{e51}.
\bibitem[{Podlubny(1998)}]{podlubny1998fractional}
\bibinfo{author}{I.~Podlubny}, \bibinfo{title}{Fractional differential
  equations: an introduction to fractional derivatives, fractional differential
  equations, to methods of their solution and some of their applications},
  volume \bibinfo{volume}{198}, \bibinfo{publisher}{Academic press},
  \bibinfo{year}{1998}.
\bibitem[{Poustelnikova et~al.(2004)Poustelnikova, Pisarev, Blagov, Samsonova,
  and Reinitz}]{poustelnikova2004database}
\bibinfo{author}{E.~Poustelnikova}, \bibinfo{author}{A.~Pisarev},
  \bibinfo{author}{M.~Blagov}, \bibinfo{author}{M.~Samsonova},
  \bibinfo{author}{J.~Reinitz},
\newblock \bibinfo{title}{A database for management of gene expression data in
  situ},
\newblock \bibinfo{journal}{Bioinformatics} \bibinfo{volume}{20}
  (\bibinfo{year}{2004}) \bibinfo{pages}{2212--2221}.
\bibitem[{Le~Gratiet(2013)}]{le2013multi}
\bibinfo{author}{L.~Le~Gratiet}, \bibinfo{title}{Multi-fidelity Gaussian
  process regression for computer experiments}, Ph.D. thesis, Universit{\'e}
  Paris-Diderot-Paris VII, \bibinfo{year}{2013}.
\bibitem[{Osborne et~al.(2008)Osborne, Roberts, Rogers, Ramchurn, and
  Jennings}]{osborne2008towards}
\bibinfo{author}{M.~A. Osborne}, \bibinfo{author}{S.~J. Roberts},
  \bibinfo{author}{A.~Rogers}, \bibinfo{author}{S.~D. Ramchurn},
  \bibinfo{author}{N.~R. Jennings},
\newblock \bibinfo{title}{Towards real-time information processing of sensor
  network data using computationally efficient multi-output gaussian
  processes},
\newblock in: \bibinfo{booktitle}{Proceedings of the 7th international
  conference on Information processing in sensor networks},
  \bibinfo{organization}{IEEE Computer Society}, pp. \bibinfo{pages}{109--120}.
\bibitem[{Zwanzig(1960)}]{zwanzig1960ensemble}
\bibinfo{author}{R.~Zwanzig},
\newblock \bibinfo{title}{Ensemble method in the theory of irreversibility},
\newblock \bibinfo{journal}{The Journal of Chemical Physics}
  \bibinfo{volume}{33} (\bibinfo{year}{1960}) \bibinfo{pages}{1338--1341}.
\bibitem[{Chorin et~al.(2000)Chorin, Hald, and Kupferman}]{chorin2000optimal}
\bibinfo{author}{A.~J. Chorin}, \bibinfo{author}{O.~H. Hald},
  \bibinfo{author}{R.~Kupferman},
\newblock \bibinfo{title}{Optimal prediction and the mori--zwanzig
  representation of irreversible processes},
\newblock \bibinfo{journal}{Proceedings of the National Academy of Sciences}
  \bibinfo{volume}{97} (\bibinfo{year}{2000}) \bibinfo{pages}{2968--2973}.

\end{thebibliography}







\end{document}